\documentclass[twoside,11pt]{article}

\usepackage{graphicx}
\usepackage{psfrag}
\usepackage{subfigure}
\usepackage{url}
\usepackage{stfloats}
\usepackage{amssymb,amsmath}
\usepackage{amsfonts}
\usepackage{listings}
\usepackage{hyperref}

\usepackage{jmlr2e}
\usepackage[multiple]{footmisc}

\jmlrheading{1}{2012}{1-4}{4/00}{10/00}{Jia Zeng}

\ShortHeadings{A Topic Modeling Toolbox Using Belief Propagation}{Jia Zeng}
\firstpageno{1}

\begin{document}

\title{A Topic Modeling Toolbox Using Belief Propagation}

\author{\name Jia Zeng \email j.zeng@ieee.org \\
       \addr School of Computer Science and Technology \\
       Soochow University\\
       Suzhou 215006, China
       }

\editor{}

\maketitle

\begin{abstract}
Latent Dirichlet allocation (LDA) is an important hierarchical Bayesian model for probabilistic topic modeling,
which attracts worldwide interests and touches on many important applications in text mining, computer vision and computational biology.
This paper introduces a topic modeling toolbox (TMBP) based on the belief propagation (BP) algorithms.
TMBP toolbox is implemented by MEX C++/Matlab/Octave for either Windows 7 or Linux.
Compared with existing topic modeling packages,
the novelty of this toolbox lies in the BP algorithms for learning LDA-based topic models.
The current version includes BP algorithms for
latent Dirichlet allocation (LDA),
author-topic models (ATM),
relational topic models (RTM),
and labeled LDA (LaLDA).
This toolbox is an ongoing project and more BP-based algorithms for various topic models will be added in the near future.
Interested users may also extend BP algorithms for learning more complicated topic models.
The source codes are freely available under the GNU General Public Licence,
Version $1.0$ at
{\em https://mloss.org/software/view/399/}.
\end{abstract}


\section{Introduction}

The past decade has seen rapid development of latent Dirichlet allocation (LDA)~\citep{Blei:03} for solving topic modeling problems
because of its elegant three-layer graphical representation as well as two efficient approximate inference methods such as
Variational Bayes (VB)~\citep{Blei:03} and collapsed Gibbs Sampling (GS)~\citep{Griffiths:04}.
Both VB and GS have been widely used to learn variants of LDA-based topic models until our recent work~\citep{Zeng:11}
reveals that there is yet another learning algorithm for LDA based on loopy belief propagation (BP).
The basic idea of BP is inspired by the collapsed GS algorithm,
in which the three-layer LDA can be interpreted as being collapsed into a two-layer Markov random field (MRF)
represented by a factor graph~\citep{Kschischang:01}.
The BP algorithm such as the sum-product operates well on the factor graph~\citep{Bishop:book}.
Extensive experiments confirm that BP is faster and more accurate than both VB and GS,
and thus is a strong candidate for becoming the standard topic modeling algorithm.
For example,
we show how to learn three typical variants of LDA-based topic models,
such as author-topic models (ATM)~\citep{Rosen-Zvi:04},
relational topic models (RTM)~\citep{Chang:10},
and labeled LDA (LaLDA)~\citep{Ramage:09}
using BP based on the novel factor graph representations~\citep{Zeng:11}.

We have implemented the topic modeling toolbox called TMBP by MEX C++ in the Matlab/Octave interface based on VB, GS and BP algorithms.
Compared with other topic modeling packages,\footnote{\url{http://www.cs.princeton.edu/~blei/lda-c/index.html}}\footnote{\url{http://psiexp.ss.uci.edu/research/programs_data/toolbox.htm},}\footnote{\url{http://nlp.stanford.edu/software/tmt/tmt-0.3/}}\footnote{\url{http://CRAN.R-project.org/package=lda}}\footnote{\url{http://mallet.cs.umass.edu/}}\footnote{\url{http://www.arbylon.net/projects/}}
the novelty of this toolbox lies in the BP algorithms for topic modeling.
This paper introduces how to use this toolbox for basic topic modeling tasks.

\section{Belief Propagation for Topic Modeling}

Given a document-word matrix $\mathbf{x} = \{x_{w,d}\}$ ($x_{w,d}$ is the number of word counts at the index $\{w,d\}$)
with word indices $1 \le w \le W$ in the vocabulary and document indices $1 \le d \le D$ in the corpus,
the probabilistic topic modeling task is to allocate topic labels $\mathbf{z} = \{z^k_{w,d}\}, z^k_{w,d} \in \{0,1\}, \sum_{k=1}^K z^k_{w,d} = 1, 1 \le k \le K$
to partition the nonzero elements $x_{w,d} \ne 0$
into $K$ topics (provided by the user) according to three topic modeling rules:
\begin{enumerate}
\item
Co-occurrence: the different word indices $w$ in the same document $d$ tend to have the same topic label.
\item
Smoothness: the same word index $w$ in the different documents $d$ tend to have the same topic label.
\item
Clustering: all word indices $w$ do not tend to be associated with the same topic label.
\end{enumerate}
Based on the above rules,
recent approximate inference methods compute the marginal distribution of topic label $\mu_{w,d}(k) = p(z^k_{w,d}=1)$ called {\em message},
and estimate parameters using the iterative EM~\citep{Dempster:77} algorithm according to the maximum-likelihood criterion.
The major difference among these inference methods lies in the message update equation.
VB updates messages by complicated digamma functions,
which cause bias and slow down message updating~\citep{Zeng:11}.
GS updates messages by topic labels randomly sampled from the message in the previous iteration.
The sampling process does not keep all uncertainty encoded in the previous message.
In contrast,
BP directly uses the previous message to update the current message without sampling.
Similar ideas have also been proposed within the approximate mean-field framework~\citep{Asuncion:10}
as the zero-order approximation of collapsed VB (CVB0) algorithm~\citep{Asuncion:09}.
While proper settings of hyperparameters can make the topic modeling performance comparable among different inference methods~\citep{Asuncion:09},
we still advocate the BP algorithms because of their ease of use and fast speed.
Table~\ref{inference} compares the message update equations among VB, GS and BP.
Compared with BP,
VB uses the digamma function $\Psi$ in message update,
and GS uses the discrete count of sampled topic labels $n^{-i}_{w,d}$ based on word tokens rather than word index in message update.
The Dirichlet hyperparameters $\alpha$ and $\beta$ can be viewed as the pseudo-messages.
The notations $-w$ and $-d$ denote all word indices except $w$ and all document indices except $d$,
and $-i$ denotes all word tokens except the current word token $i$.
More details can be found in our work~\citep{Zeng:11,Zeng:12a}.

\begin{table}[t]
\centering
\caption{Comparison of message update equations~\citep{Zeng:11}.}
\begin{tabular}{|c|c|} \hline
Inference methods   &Message update equations            \\ \hline \hline
VB                  &$\mu_{w,d}(k) \propto \frac{\exp[\Psi(x_{\cdot,d}\mu_{\cdot,d}(k) + \alpha)]}{\exp[\Psi(\sum_k [x_{\cdot,d}\mu_{\cdot,d}(k) + \alpha])]} \times
\frac{x_{w,\cdot}\mu_{w,\cdot}(k) + \beta}{\sum_w [x_{w,\cdot}\mu_{w,\cdot}(k) + \beta]}$      \\ \hline
GS                  &$\mu_{w,d,i}(k) \propto \frac{n^{-i}_{\cdot,d}(k) + \alpha}{\sum_k [n^{-i}_{\cdot,d}(k) + \alpha]} \times
\frac{n^{-i}_{w,\cdot}(k) + \beta}{\sum_w [n^{-i}_{w,\cdot}(k) + \beta]}$      \\ \hline
BP                  &$\mu_{w,d}(k) \propto \frac{x_{-w,d}\mu_{-w,d}(k) + \alpha}{\sum_k [x_{-w,d}\mu_{-w,d}(k) + \alpha]} \times
\frac{x_{w,-d}\mu_{w,-d}(k) + \beta}{\sum_w [x_{w,-d}\mu_{w,-d}(k) + \beta]}$ \\ \hline
\end{tabular}
\label{inference}
\end{table}

Because VB and GS have been widely used for learning different LDA-based topic models,
it is easy to develop the corresponding BP algorithms for learning these LDA-based topic models by
either removing the digamma function in the VB or without sampling from the posterior probability in the GS algorithm.
For example,
we show how to develop the corresponding BP algorithms for two typical LDA-based topic models such as ATM and RTM~\citep{Zeng:11}.

\section{An Example of Using TMBP}

TMBP toolbox contains source codes for learning LDA based on VB, GS, and BP~\citep{Zeng:11,Zeng:12a,Zeng:12b},
learning author-topic models (ATM)~\citep{Rosen-Zvi:04} based on GS and BP,
learning relational topic models (RTM)~\citep{Chang:10} and labeled LDA~\citep{Ramage:09} using BP.
Implementation details can be found in ``readme".

Here,
we present a demo for the synchronous BP algorithm.
After installation,
we run \verb+demo1.m+ in the Octave/Matlab environment.
The results (the training perplexity at every 10 iterations and the top five words in each of ten topics) are printed on the screen:
\begin{lstlisting}
*********************
The sBP Algorithm
*********************
    Iteration 10 of 500:    1041.620873
    ...
    ...
    Iteration 490 of 500:   741.946849
Elapsed time is 13.246747 seconds.

*********************
Top five words in each of ten topics by sBP
*********************
design system reasoning case knowledge
model models bayesian data markov
genetic problem search algorithms programming
algorithm learning number function model
learning paper theory knowledge examples
learning control reinforcement paper state
model visual recognition system patterns
research report technical grant university
network neural networks learning input
data decision training algorithm classification
\end{lstlisting}


\acks{This work is supported by NSFC (Grant No. 61003154),
the Shanghai Key Laboratory of Intelligent Information Processing,
China (Grant No. IIPL-2010-009),
and a grant from Baidu.}


%

\vskip 0.2in
\bibliographystyle{natbib}
\bibliography{IEEEabrv,BPTM}

\begin{thebibliography}{13}
\providecommand{\natexlab}[1]{#1}
\providecommand{\url}[1]{\texttt{#1}}
\expandafter\ifx\csname urlstyle\endcsname\relax
  \providecommand{\doi}[1]{doi: #1}\else
  \providecommand{\doi}{doi: \begingroup \urlstyle{rm}\Url}\fi

\bibitem[Asuncion(2010)]{Asuncion:10}
Arthur Asuncion.
\newblock {Approximate Mean Field for Dirichlet-Based Models}.
\newblock In \emph{ICML Workshop on Topic Models}, 2010.

\bibitem[Asuncion et~al.(2009)Asuncion, Welling, Smyth, and Teh]{Asuncion:09}
Arthur Asuncion, Max Welling, Padhraic Smyth, and Yee~Whye Teh.
\newblock {On smoothing and inference for topic models}.
\newblock In \emph{UAI}, pages 27--34, 2009.

\bibitem[Bishop(2006)]{Bishop:book}
C.~M. Bishop.
\newblock \emph{Pattern recognition and machine learning}.
\newblock Springer, 2006.

\bibitem[Blei et~al.(2003)Blei, Ng, and Jordan]{Blei:03}
D.~M. Blei, A.~Y. Ng, and M.~I. Jordan.
\newblock Latent {Dirichlet} allocation.
\newblock \emph{J. Mach. Learn. Res.}, 3:\penalty0 993--1022, 2003.

\bibitem[Chang and Blei(2010)]{Chang:10}
J.~Chang and D.~M. Blei.
\newblock Hierarchical relational models for document networks.
\newblock \emph{Annals of Applied Statistics}, 4\penalty0 (1):\penalty0
  124--150, 2010.

\bibitem[Dempster et~al.(1977)Dempster, Laird, and Rubin]{Dempster:77}
A.~P. Dempster, N.~M. Laird, and D.~B. Rubin.
\newblock {Maximum likelihood from incomplete data via the EM algorithm}.
\newblock \emph{Journal of the Royal Statistical Society, Series B},
  39:\penalty0 1--38, 1977.

\bibitem[Griffiths and Steyvers(2004)]{Griffiths:04}
T.~L. Griffiths and M.~Steyvers.
\newblock Finding scientific topics.
\newblock \emph{Proc. Natl. Acad. Sci.}, 101:\penalty0 5228--5235, 2004.

\bibitem[Kschischang et~al.(2001)Kschischang, Frey, and
  Loeliger]{Kschischang:01}
F.~R. Kschischang, B.~J. Frey, and H.-A. Loeliger.
\newblock Factor graphs and the sum-product algorithm.
\newblock \emph{IEEE Transactions on Inform. Theory}, 47\penalty0 (2):\penalty0
  498--519, 2001.

\bibitem[Ramage et~al.(2009)Ramage, Hall, Nallapati, and Manning]{Ramage:09}
Daniel Ramage, David Hall, Ramesh Nallapati, and Christopher~D. Manning.
\newblock {Labeled LDA: A supervised topic model for credit attribution in
  multi-labeled corpora}.
\newblock In \emph{Empirical Methods in Natural Language Processing}, pages
  248--256, 2009.

\bibitem[Rosen-Zvi et~al.(2004)Rosen-Zvi, Griffiths, Steyvers, and
  Smyth]{Rosen-Zvi:04}
M.~Rosen-Zvi, T.~Griffiths, M.~Steyvers, and P.~Smyth.
\newblock The author-topic model for authors and documents.
\newblock In \emph{UAI}, pages 487--494, 2004.

\bibitem[Zeng et~al.(2011)Zeng, Cheung, and Liu]{Zeng:11}
Jia Zeng, William~K. Cheung, and Jiming Liu.
\newblock Learning topic models by belief propagation.
\newblock \emph{IEEE Trans. Pattern Anal. Mach. Intell.}, page
  arXiv:1109.3437v4 [cs.LG], 2011.

\bibitem[Zeng et~al.(2012{\natexlab{a}})Zeng, Liu, and Cao]{Zeng:12a}
Jia Zeng, Zhi-Qiang Liu, and Xiao-Qin Cao.
\newblock A new approch to speeding up topic modeling.
\newblock \emph{IEEE Trans. Pattern Anal. Mach. Intell.}, page
  arXiv:1204.0170v1 [cs.LG], 2012{\natexlab{a}}.

\bibitem[Zeng et~al.(2012{\natexlab{b}})Zeng, Liu, and Cao]{Zeng:12b}
Jia Zeng, Zhi-Qiang Liu, and Xiao-Qin Cao.
\newblock Tiny belief propagation for training latent {Dirichlet} allocation
  with less memeory.
\newblock \emph{J. Mach. Learn. Res.}, page submitted, 2012{\natexlab{b}}.

\end{thebibliography}

\end{document}